\documentclass{article}

\usepackage{PRIMEarxiv}

\usepackage[utf8]{inputenc} % allow utf-8 input
\usepackage[T1]{fontenc}    % use 8-bit T1 fonts
\usepackage{hyperref}       % hyperlinks
\usepackage{url}            % simple URL typesetting
\usepackage{booktabs}       % professional-quality tables
\usepackage{amsfonts}       % blackboard math symbols
\usepackage{nicefrac}       % compact symbols for 1/2, etc.
\usepackage{microtype}      % microtypography
\usepackage{lipsum}
\usepackage{fancyhdr}       % header
\usepackage{graphicx}       % graphics
\graphicspath{{media/}}     % organize your images and other figures under media/ folder

\usepackage{amssymb}
\usepackage{latexsym}
\usepackage{amsmath}
\usepackage{amsthm}
\usepackage{adjustbox}

\usepackage{algorithm,algorithmic}

\theoremstyle{plain}
\newtheorem{theorem}{Theorem}[section]

\newtheorem{lemma}[theorem]{Lemma}
\newtheorem{corollary}[theorem]{Corollary}
\theoremstyle{definition}
\newtheorem{definition}[theorem]{Definition}

\theoremstyle{remark}
\newtheorem{remark}[theorem]{Remark}

\title{CP-PINNs: Data-Driven Changepoints Detection in PDEs Using Online Optimized Physics-Informed Neural Networks
\thanks{Under consideration at Pattern Recognition Letters.} 
}

\author{
  \large Zhikang Dong$^{a}$ \quad \quad \quad Pawe\l \ Polak$^{a\, b\, c}$ \\
 \\
 \\
 \normalsize
  $^{a}$Department of Applied Mathematics and Statistics\\
  \normalsize
  $^{b}$Institute for Advanced Computational Science\\
  \normalsize
  $^{c}$Center of Excellence in Wireless and Information Technology\\
  Stony Brook University\\
  Stony Brook, NY, 11794, USA \\
  \texttt{\{zhikang.dong.1, pawel.polak\}@stonybrook.edu}
}

\begin{document}
\maketitle

\begin{abstract}
We investigate the inverse problem for Partial Differential Equations (PDEs) in scenarios where the parameters of the given PDE dynamics may exhibit changepoints at random time. We employ Physics-Informed Neural Networks (PINNs) – universal approximators capable of estimating the solution of any physical law described by a system of PDEs, which serves as a regularization during neural network training, restricting the space of admissible solutions and enhancing function approximation accuracy. We demonstrate that when the system exhibits sudden changes in the PDE dynamics, this regularization is either insufficient to accurately estimate the true dynamics, or it may result in model miscalibration and failure. Consequently, we propose a PINNs extension using a Total-Variation penalty, which allows to accommodate multiple changepoints in the PDE dynamics and significantly improves function approximation. These changepoints can occur at random locations over time and are estimated concurrently with the solutions. Additionally, we introduce an online learning method for re-weighting loss function terms dynamically. Through empirical analysis using examples of various equations with parameter changes, we showcase the advantages of our proposed model. In the absence of changepoints, the model reverts to the original PINNs model. However, when changepoints are present, our approach yields superior parameter estimation, improved model fitting, and reduced training error compared to the original PINNs model.
\end{abstract}

% keywords can be removed
% \keywords{First keyword \and Second keyword \and More}

\section{Introduction}
\label{intro}

Deep learning and machine learning methods are widely studied and used in academia and industry. They perform successfully in tasks such as dimensionality reduction \cite{kingma2013auto}, computer vision \cite{szegedy2016rethinking,kim2024face}, multimodal learning \cite{dong2023musechat,liu2024tackling}, and time series analysis \cite{greff2016lstm}. Recent advancements have further expanded the applicability of deep learning models, demonstrating their capability to approximate a broad spectrum of nonlinear functions within complex dynamical systems \cite{mhaskar2019function,lusch2018deep,daubechies2022nonlinear}. A key advancement is the advent of Physics-Informed Neural Networks (PINNs), which combines the adaptability of neural networks with the physics of Partial Differential Equations (PDEs) and data-driven techniques for forward and inverse problems \cite{raissi2019physics}.  Unlike conventional numerical methods \cite{bar1999fitting,muller2002fitting,xun2013parameter}, PINNs integrate supervised learning tasks with the enforcement of physical laws, represented by general nonlinear PDEs, serving as prior information.

There are many variations of PINNs, e.g., Physics-informed generative adversarial networks \cite{yang2020physics} which have stochastic differential equations induced generators to tackle very high dimensional problems; \cite{han2018solving} rewrites PDEs as backward stochastic differential equations and designs the gradient of the solution as policy function, which is approximated by deep neural networks. In reinforcement learning, \cite{yang2021b} proposes a Bayesian neural network as the prior for PDEs and use Hamiltonian Monte Carlo and variational inference as the estimator of the posterior, resulting in more accurate prediction and less overfitting. Based on Petrov-Galerkin method, the hp-variational PINNs (hp-VPINNs) \cite{kharazmi2021hp} allows for localized parameters estimation with given test functions via domain decomposition. The hp-VPINNs generates a global approximation to the weak solution of the PDE with local learning algorithm that uses a domain decomposition which is preselected manually.

Existing PINNs methods face challenges in managing abrupt variations or discontinuities in dynamical systems. Such changes often signal shifts in system dynamics or the influence of external factors. For example, detecting leakages in pipelines using limited sensor data \cite{aamo2015leak}; traffic flow management by predicting congestion without comprehensive sensor coverage \cite{laval2013hamilton}; environmental monitoring for sudden pollutant concentration shifts \cite{wang2018prediction}; ensuring energy grid stability by pinpointing fluctuations or failures \cite{yang2019data,tartakovsky2014rapid}; overseeing heat distribution in manufacturing materials \cite{zobeiry2021physics}; medical imaging to detect tissue property changes \cite{oliphant2001complex}; seismic activity monitoring for early earthquake detection \cite{haghighat2023novel}; aerospace component stress assessment for safety \cite{singh2021non}; agriculture water and nutrient distribution tracking \cite{bao2021partial,song2021modeling}; atmospheric change detection in weather systems \cite{muller2014massively}; and ensuring product consistency in manufacturing \cite{van2007modeling}. 

While changepoints detection methods have shown promise in identifying significant shifts in data characteristics across various fields—from high-dimensional time series data \cite{safikhani2022joint,bai2020multiple,balabhadra2023high}, computer vision \cite{radke2005image,li2024feature}, speech recognition \cite{chowdhury2012bayesian,rybach2009audio}, real-time medical monitoring \cite{staudacher2005new,yang2006adaptive}, to disturbance localization in power grids \cite{yang2019data}, and anomaly detection in computer networks \cite{tartakovsky2014rapid}—their integration into PINNs for detecting dynamical changepoints in PDEs remains unexplored. This gap underscores a critical need for developing methodologies that can adapt changepoints detection techniques to the specific challenges of PDE dynamics.

Online learning methods enable model updates incrementally from sequential data, offering greater efficiency and scalability than traditional batch learning.  Regularization technique is widely used in online convex optimization problems \cite{miao2023online}. Online Mirror Descent, an extension of Mirror Descent \cite{nemirovskij1983problem}, utilizes a gradient update rule in the dual space, leading to improved bounds. Adaptive subgradient method \cite{duchi2011adaptive} dynamically adjusts regularization term based on its current subgradient. Follow-the-Regularized-Leader \cite{abernethy2008competing,shalev2007primal} is stable extension of Follow-the-Leader \cite{kalai2005efficient,hannan1957approximation} by adding a strong convex regularization term to the objective function to achieve a sublinear regret bound.

In this work, we present an innovative methodology that combines changepoints detection with PINNs to address changes and instabilities in the dynamics of PDEs. This approach marks the first exploration into simultaneously detecting changepoints and estimating unknown parameters within PDE dynamics based on observed data. We have three main contributions: (i) We introduce a novel strategy that leverages PINNs alongside the Total Variation method for detecting changepoints within PDE dynamics. This approach not only identifies the timing of changes but also facilitates the estimation of unknown system parameters. (ii) We propose an online learning technique aimed at optimizing the weights within the loss function during training. By adaptively adjusting these weights, our method not only enhances the model's estimation accuracy but also increases its robustness against the instabilities associated with rapid parameter variations. (iii) We present several theoretical results to show that our re-weighting approach minimizes the training loss function with a regularizer and demonstrates that the regret is upper bounded. The theoretical results also indicate that the weight update method does not alter the neural network's optimization objective on average.
\section{Model and Estimation}\label{sec:Model} 
We consider a general form of parameterized and nonlinear continuous partial equations with changepoints
\begin{equation} \label{def2.1}
u_{t}+\mathcal{N}[u ; \lambda(t)]=0,
\end{equation}
where $(\mathbf{x}, t) \in \Omega \times(0, T]$, $\Omega \subset \mathbb{R}^{d}$ is the bounded domain, $\mathcal{N}[\cdot ; \lambda(t)]$ is a nonlinear operator parameterized by a function $\lambda(t)$, and  $u(\mathbf{x}, t)$ is the latent solution to the above equation.
\begin{definition}\label{ass:piecewise_lambda}
The function $\lambda(t):[0,T]\to\mathbb{R}$ is piecewise-constant, bounded, with a finite but unknown number of discontinuities $\kappa^*$, located at some of the observations, i.e., $\lambda(t) = \sum_{i=1}^{\kappa^*} \lambda_i \mathbf{I}_{[t_{i-1},t_i)}(t)$, for $0=t_0<t_1<\ldots<t_{\kappa^*}<T$, where $\mathbf{I}_{A}(t)$ is the indicator function of set $A$.
\end{definition}
\subsection{Physics-Informed Neural Network}
Following \cite{raissi2019physics}, in order to approximate the true solution of \eqref{def2.1}, we use the neural network approximation $u_{N N}(\mathbf{x}, t ; \mathbf{\Theta})$ given by
\begin{equation}\label{eq:u_nn_def}
u_{N N}\left(\mathbf{x}, t ; \mathbf{\Theta}\right)=g \circ T^{(\ell)} \circ T^{(\ell-1)} \circ \cdots \circ T^{(1)}([\mathbf{x},t]).
\end{equation}
For each hidden layer $i=1,\ldots \ell$, the nonlinear operator $T^{(i)}$ is defined as $T^{(i)}([\mathbf{x},t])=\sigma\left(\mathbf{W}_{i} [\mathbf{x},t]+\mathbf{b}_{i}\right)$ with weights $\mathbf{W}_{i} \in \mathbb{R}^{\mathcal{M}_{i} \times \mathcal{M}_{i-1}}$ and biases $\mathbf{b}_{i} \in \mathbb{R}^{\mathcal{M}_{i}}$, where $\mathcal{M}_{0}=d$ is the input dimension and in the output layer, the operator $g: \mathbb{R}^{M_{\ell}} \rightarrow \mathbb{R}$ is a linear activation function. Next, we can define a feature function with respect to \eqref{def2.1} and approximate $u_t$ using deep neural networks.
\begin{equation} 
f(\mathbf{x}, t)=u_{t}+\mathcal{N}[u ; \lambda (t)].
\end{equation}
We write the corresponding boundary and initial residual functions
\begin{equation} \label{residual}
\begin{aligned}
r_{b}(u_{NN}, \mathbf{x}, t)&=u_{NN}-u, & \forall(\mathbf{x}, t) \in \partial \Omega \times(0, T]\\
r_{0}(u_{NN}, \mathbf{x}, t)&=u_{NN}-u_0, & \forall(\mathbf{x}, t) \in \Omega \times\{t=0\}. 
\end{aligned}\end{equation}
Averaging over observations on their domains, we obtain the fitting loss function for the training of the model
\begin{equation} \label{f_loss}
L^{\mathfrak{f}}=\frac{1}{N_{b}} \sum_{i=1}^{N_{b}}\left|r_{b}\left(\mathbf{x}_{b}^{i}, t_{b}^{i}\right)\right|^{2}+\frac{1}{N_{0}} \sum_{i=1}^{N_{0}}\left|r_{0}\left(\mathbf{x}_{0}^{i}\right)\right|^{2}.
\end{equation}
Similarly, for the observations inside the domain, the structure loss is defined as
\begin{equation} \label{s_loss}
L^{\mathfrak{s}}=\frac{1}{N} \sum_{i=1}^{N}\left|f_{NN}(\mathbf{x}^i, t^i)\right|^{2}.
\end{equation}
By gathering together these two loss functions with equal weights, one obtains the total loss for the training of the original PINNs model as defined in  \cite{raissi2019physics}. 

\subsection{Total Variation Regularization}

The standard PINNs model assumes that the parameters of PDEs are constant values across the entire time domain. In order to accommodate Definition \ref{ass:piecewise_lambda}, we allow for the changes in the $\lambda(t)$ and introduce additional regularization term in a form of total variation penalty on the first difference in $\lambda(t)$.
\begin{equation} \label{tvr}
V^{\mathfrak{{\lambda}}} = \sum_{i=1}^{T-1} \delta(t^i)\left|\Delta{\lambda}(t^i)\right|,
\end{equation}
where $\Delta{\lambda}(t^i) = {\lambda}(t^{i+1}) - {\lambda}(t^i)$, and $\delta(t)$ is a U-shape linear function of $t$ which increases the penalty strength closer to the edges of time domain in order to avoid estimation instabilities---in the experiments below we use $\delta(t) = \sqrt{T/(T-t)}$. The  $V^{\mathfrak{{\lambda}}}$ in \eqref{tvr} is a sparsity inducing penalty similar to one used in \cite{harchaoui2010multiple,tibshirani-adaptive} for changepoints detection in linear regression and non-parametric statistics, respectively. After estimating the changepoints locations, we perform the standard PINNs to refine PDEs discovery and solve PDEs within each detected time interval.

In order to avoid non-differentiability of the objective function and due to its extensive optimization and prevalent usage in deep learning frameworks such as PyTorch and TensorFlow, we use the ReLU activation function in expressing $V^{\mathfrak{\lambda}}$. We have
\begin{equation}\label{eq:differentiable_V}
V^{\mathfrak{\lambda}} = \sum_{i=1}^{T-1} \delta(t^i)\left[\Delta\lambda^+(t^i) + \Delta\lambda^-(t^i)\right],
\end{equation}
where $\Delta\lambda^+(t^i) = \max(\Delta\lambda(t^i),0)$ and 
$\Delta\lambda^-(t^i) = \max(-\Delta\lambda(t^i),0)$.
Then our optimizer runs in terms $\Delta\lambda^+(t^i)$ and $\Delta\lambda^+(t^i)$ subject to $\Delta\lambda(t^i) =\Delta\lambda^+(t^i) + \Delta\lambda^-(t^i)$.

\subsection{Exponentiated Descent Weights Update Method for Loss Function}
In order to balance the three goals during the training process, we define our loss function as a weighted average of the loss terms
\begin{equation} \label{nondiff_total_loss}
L(\mathbf{w};\boldsymbol{\Theta},\lambda(t)) = \left[L^{\mathfrak{f}}, L^{\mathfrak{s}}, V^{\mathfrak{{\lambda}}}\right]^\top\mathbf{w},
\end{equation}
where $L^{\mathfrak{f}}>0$, $L^{\mathfrak{s}}>0$, $V^{\mathfrak{{\lambda}}}>0$, $\mathbf{w}\in \mathcal{S}_3$ is the vector of weights towards corresponding loss terms, and $\mathcal{S}_3 = \left\{\mathbf{w}=[w_1,w_2,w_3]^\top \in\mathbb{R}^3: \mathbf{w}^\top \mathbf{1}_3=1 \text{ and } w_i>0\right\}$, where $\mathbf{1}_3=[1,1,1]^\top$.

\begin{figure}[t]
\centering
\includegraphics[height=4.5cm, width=1\linewidth]{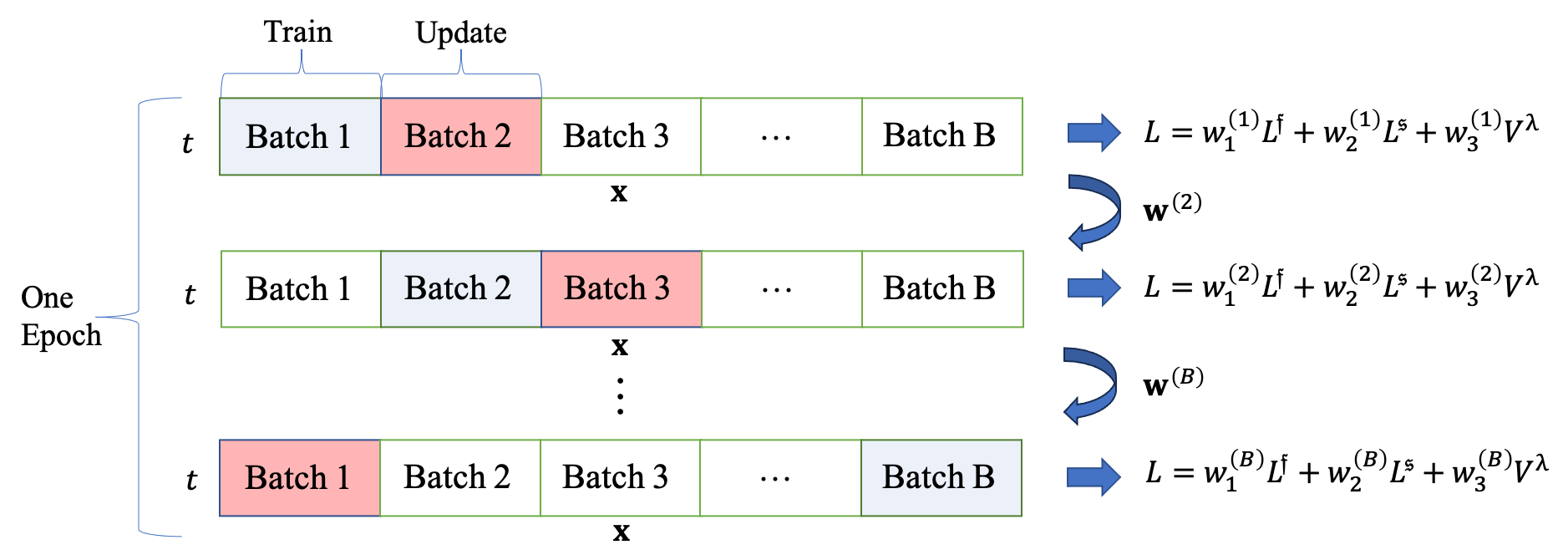} 
\caption{The dataset is partitioned based on spatial information, with each batch encompassing the full temporal information. In the online learning approach, the network is trained using the previous distribution of loss weights and updated based on the data from the subsequent batch.}
\label{fig:flowchart}
\end{figure}

Our goal is to find $\mathbf{w}$ which minimizes the loss function in \eqref{nondiff_total_loss} on the next batch of data (``Update'' in Figure \ref{fig:flowchart}). However, directly minimizing loss function may lead to a very large change in $\mathbf{w}$ and lead to instability during the training. Therefore, we minimize the loss function with a negative-entropy regularization term:
\begin{equation}\label{eq:objective_func}
\mathbf{w}^{(k)}=\underset{\mathbf{w}\in\mathcal{S}_3}{\operatorname{argmin}} \left\{ L(\mathbf{w}) + \frac{1}{\eta}\mathbf{w}^\top\log\mathbf{w}\right\}.
\end{equation}
The following Lemma derives the \textit{exponentiated descent weights update method} (see also  \cite{mcmahan2011follow}, and  \cite[Chp. 2]{shalev2012online}) which solves \eqref{eq:objective_func}.
\begin{lemma}\label{lemma:update_weigts}
The solution to \eqref{eq:objective_func} is given by
\begin{equation}\label{eq:decision}
    \left[ 
\begin{array}{c} 
w_1^{(k)} \\ 
w_2^{(k)} \\ 
w_3^{(k)}
\end{array} 
\right] = \left[ 
\begin{array}{c} 
\exp\left[-\eta L^{\mathfrak{f}}_{(k)} - \left( 1-\eta\gamma\right)\right] \\ 
\exp\left[-\eta L^{\mathfrak{s}}_{(k)} - \left( 1-\eta\gamma\right)\right]\\ 
\exp\left[-\eta V^{\mathfrak{{\hat{\lambda}}}}_{(k-1)} - \left( 1-\eta\gamma\right)\right]
\end{array} 
\right],
\end{equation}
where 
\begin{equation}\label{eq:3_loss_terms}
    \begin{aligned}
    L^{\mathfrak{f}}_{(k)} &= \frac{1}{N}\sum_{i=1}^{N} \left( u_i - u_{NN}(\mathbf{x}^{(k)}_i, t, \mathbf{\hat{\Theta}}^{(k-1)}(\mathbf{w}^{(k-1)})\right)^2, \\
    L^{\mathfrak{s}}_{(k)} &= \frac{1}{N} \sum_{i=1}^{N}\left|f_{NN}(\mathbf{x}^{(k)}_i, t, \mathbf{\hat{\Theta}}^{(k-1)}(\mathbf{w}^{(k-1)}))\right|^{2}, \\
    V^{\mathfrak{{\hat{\lambda}}}}_{(k-1)} &= \sum_{i=1}^{T-1} \delta(i)\left[\Delta\hat{\lambda}^+_{(k-1)}(t^i) + \Delta\hat{\lambda}^-_{(k-1)}(t^i)\right],
\end{aligned}
\end{equation}
and $\gamma$ is such that the vector $\mathbf{w}^{(k)}$ is guaranteed to remain in the simplex $\mathcal{S}_3$, i.e.,
\begin{equation}\label{eq:gamma_simplex}
    \gamma = \frac{1}{\eta}\left(1 - \log\left(e^{-\eta L^{\mathfrak{f}}_{(k)}} + e^{-\eta L^{\mathfrak{s}}_{(k)}} + e^{-\eta  V^{\mathfrak{{\hat{\lambda}}}}_{(k-1)}}\right)\right).
\end{equation}
\end{lemma}
The proof for Lemma \ref{lemma:update_weigts} is in the Appendix. 
We are interested in the stability of the algorithm with adaptive weights. Therefore, we look at the difference between the loss function values with adaptive weights versus any fixed weights after $B$ batches (where $\Theta$ and $\lambda(t)$ are estimated on this batch). This can be summarized using the definition of regret as
\begin{equation}
\operatorname{Regret} = \sum_{k=1}^B L(\mathbf{w}^{(k)}; \mathbf{\hat{\Theta}}^{(k)},\hat{\lambda}^{(k)}(t))-\sum_{k=1}^B L(\mathbf{w}^{*}; \mathbf{\hat{\Theta}}^{(k)},\hat{\lambda}^{(k)}(t)).
\end{equation}
The following corollary provides the regret bound for our algorithm.
\begin{corollary}\label{cor:regret}
Let  $\|\nabla_w L(\mathbf{w}^{(k)})\|_1 < G$, where $\mathbf{w}^{k}$ solves \eqref{eq:objective_func}, then 
\begin{equation}
\operatorname{Regret} \leq \frac{\log 3}{\eta} + \eta B G^2,\quad \forall  \mathbf{w}^{*}\in \mathcal{S}_3
\end{equation}
\end{corollary}
The proof for Corollary \ref{cor:regret} is in the Appendix.
\begin{remark}\label{rem:regret_sublinear}
By choosing $\eta^* = \frac{\sqrt{\log 3}}{G\sqrt{B}}$, we get $\operatorname{Regret} \leq 2G\sqrt{B\log 3}$, $\forall  \mathbf{w}^{*}\in \mathcal{S}_3$.\end{remark}
Note that in the definition of Regret, $\mathbf{\hat{\Theta}}^{(k)}$ and $\hat{\lambda}(t)^{(k)}$ are jointly estimated on the $k$-th batch, while $\mathbf{w}^{(k)}$ is obtained using only information from the previous batches. The sublinearity in $B$ of the regret bound in Remark \ref{rem:regret_sublinear} implies that updating the weights using \eqref{eq:objective_func} on average does not impact the loss. Hence, our algorithm gradually improves the loss function using the out-of-sample information from the next ("update") batch, without negatively impacting the average stability of the neural-network training compared with the analogous loss function with fixed weights $\mathbf{w}^*\in \mathcal{S}_3$.

\begin{algorithm}[h!]
\renewcommand{\thealgorithm}{}
\begin{algorithmic}[2]
\REQUIRE The spatio information $\mathbf{x}$ and real solution $u$; the batch size $B$; the fixed learning rate $\eta>0$.
\ENSURE The target parameter function $\lambda \left(t\right)$ from Definition \ref{ass:piecewise_lambda} and estimated PDEs solution $u_{NN}$  from \eqref{eq:u_nn_def}.
% \vspace{2mm}
\\\hrulefill
\vspace{2mm}
\\ \textbf{Initialization}: Fix the learning rate $\eta$ and the starting weights $\mathbf{w}^{(0)}$.
In each epoch, \\ 
\textbf{for} $k= 1,2,\dots,B$ \textbf{do}
\begin{itemize}
    \item[1.] Estimate jointly the PDEs solution $u_{N N}$ and the target parameter $\lambda^{(k-1)}\left( t \right)$ by minimizing \eqref{nondiff_total_loss} using data from batch $k$ and weights $\mathbf{w}^{(k-1)}$.
    \item[2.] Compute the loss   and $\gamma$ defined in \eqref{eq:3_loss_terms} and \eqref{eq:gamma_simplex}, respectively.
    \item[3.] Update the weights $\mathbf{w}^{(k)}$ using \eqref{eq:decision}.
\end{itemize}
\textbf{end for}
\end{algorithmic}
\caption{Loss Function Weights Update Algorithm}
\label{al2}
\end{algorithm}
\begin{figure*}[h]
\centering
  {%
  \includegraphics[height=4.5cm,width=.49\linewidth]{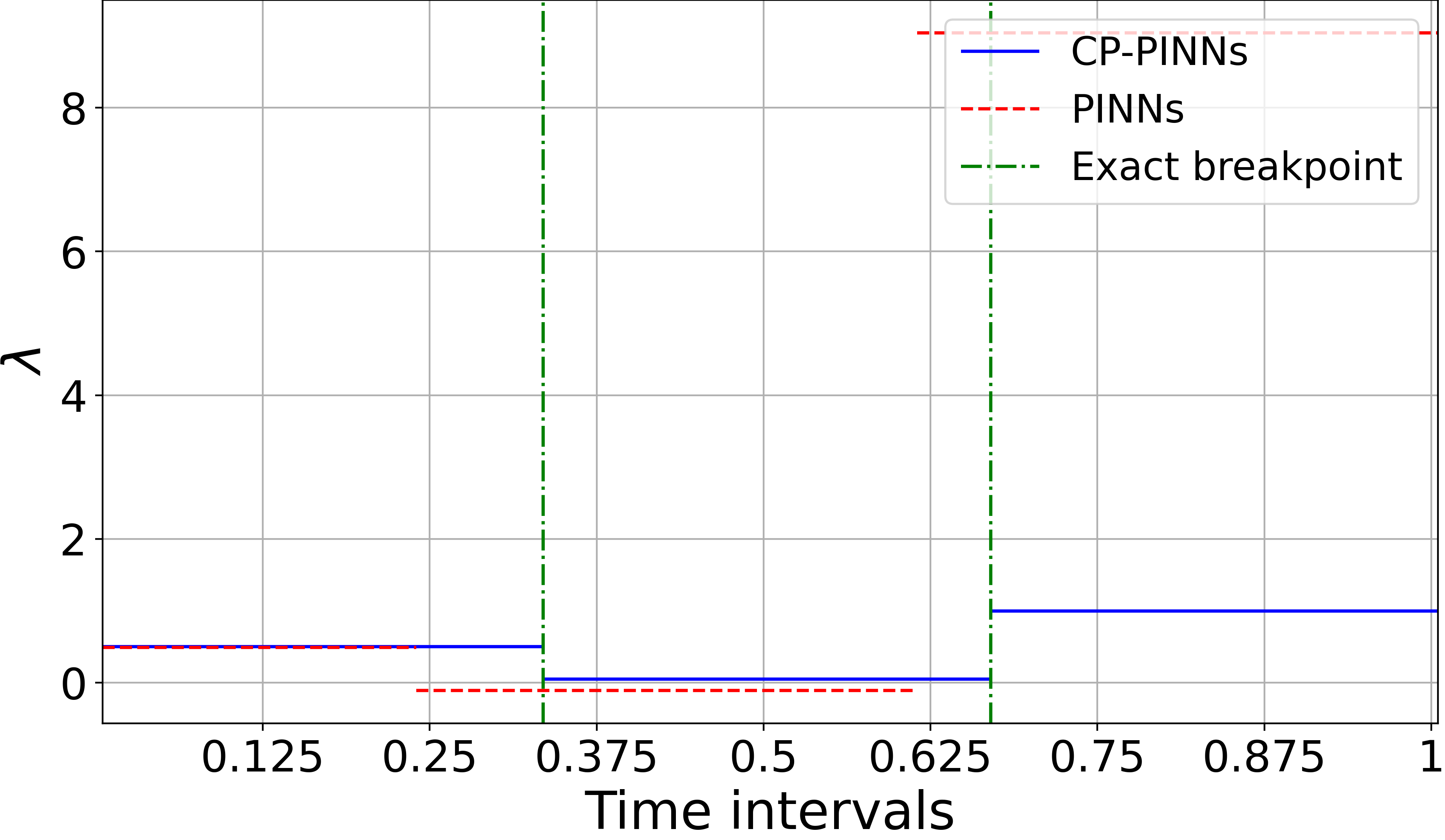}
}
  \hfill
  {%
  \includegraphics[height=4.5cm,width=.49\linewidth]{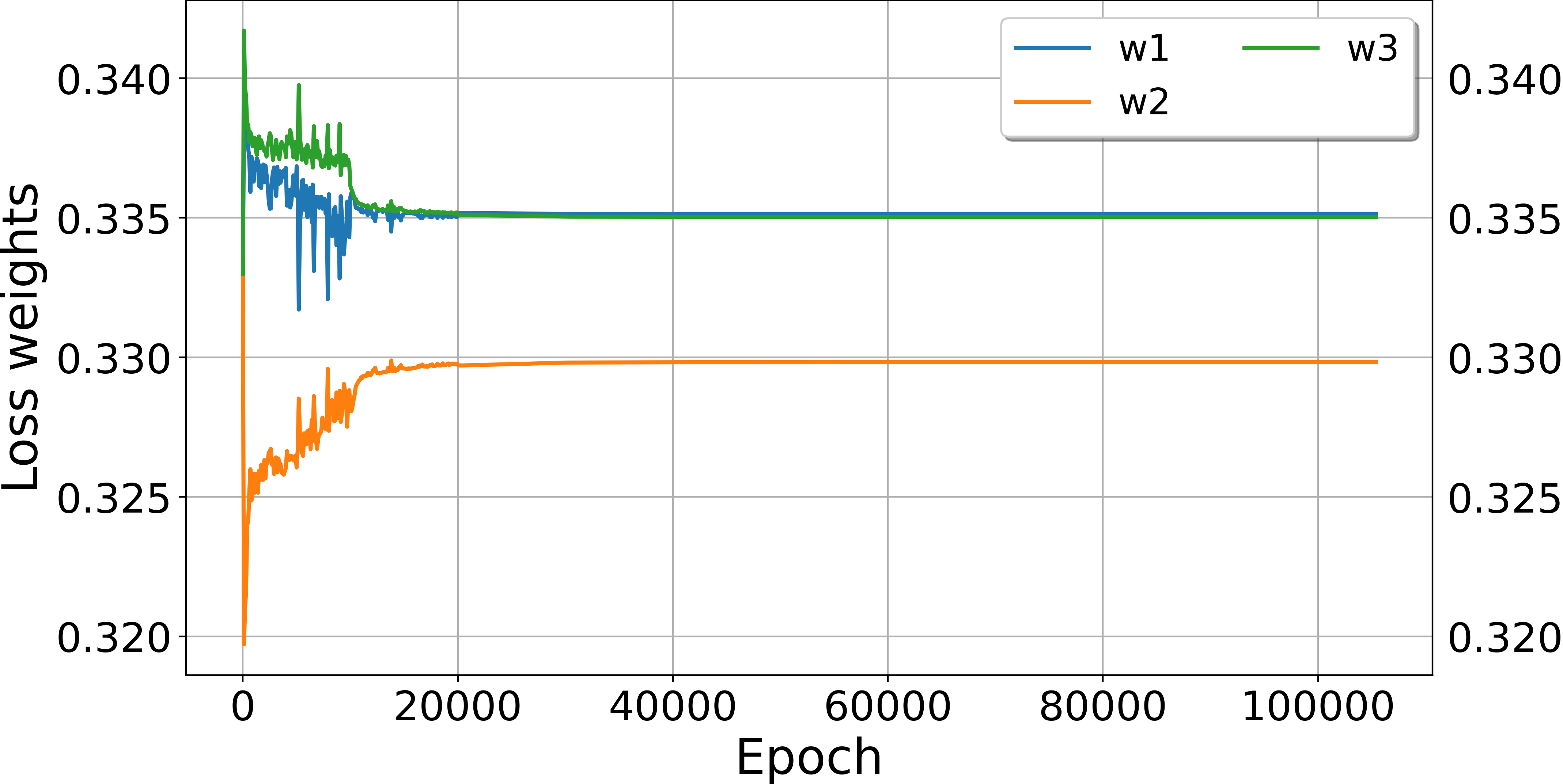}
}
\caption{(Left) Parameter estimation of CP-PINNs and PINNs. (Right) Loss weights distribution across time.}
\label{fig:2_change_point_ade}
\end{figure*}
\section{Empirical Results}\label{sec:empirics}
For our Loss Function Weights Update Algorithm, we set $\mathbf{w}^{(0)}=\left[\frac{1}{3},\frac{1}{3},\frac{1}{3}\right]^{\top}$ which reflects equal weighting of all the loss terms analogously to the original PINNs model.  For the learning rate we experimented with different $\eta>0$ values in the range of $\left[10^{-6},10^{-3}\right]$ and the algorithm in all the experiments was very stable and converged numerically to the same solution. We use $\eta=10^{-4}$ as the example here.

For the first experiment, we consider the advection-diffusion equation, which is widely used in heat transfer, mass transfer, and fluid dynamics problems. The equation of one-dimensional form is given by
\begin{equation}\label{eq:advection}
\begin{aligned}
    \frac{\partial u}{\partial t} + \frac{\partial u}{\partial x} &= \lambda(t) \frac{\partial^{2} u}{\partial x^{2}}, \\
    u(t, -1) &= u(t, 1) = 0, \\
    u(x, 0) &= -\sin (\pi x),
\end{aligned}
\end{equation}
where $\lambda(t)=0.5$ for $t\in[0,\frac{1}{3})$, $0.05$ for $t\in[\frac{1}{3},\frac{2}{3})$ and $1$ for $t\in[\frac{2}{3},1]$. $u(x, t): \Omega \rightarrow \mathbb{R}$, and $\Omega=[-1,1] \times[0,1]$.
\begin{figure*}[h!]
\centering
{%
  \includegraphics[height=4.5cm,width=.25\linewidth]{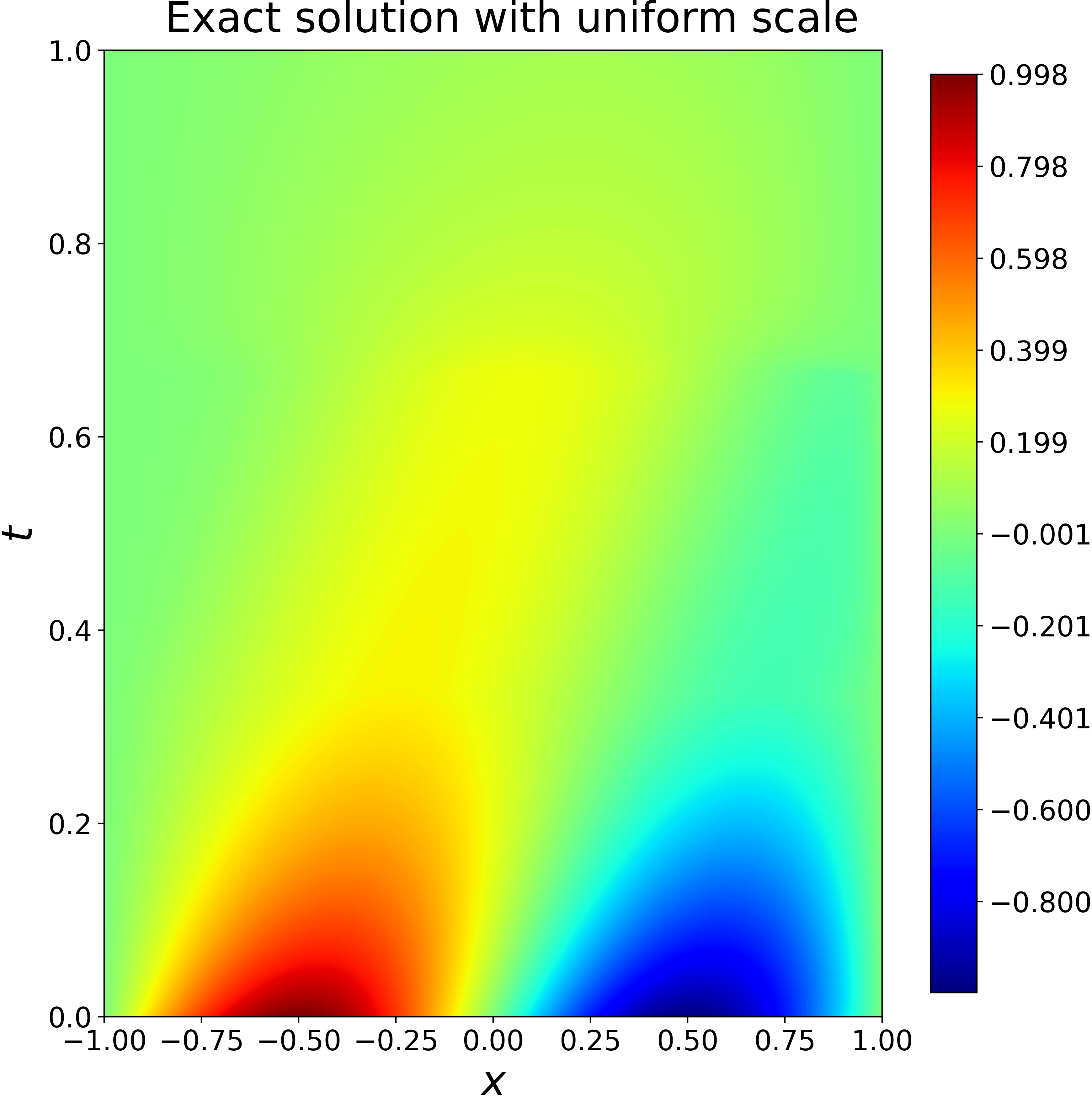}%
}\hfill{%
  \includegraphics[height=4.5cm,width=.25\linewidth]{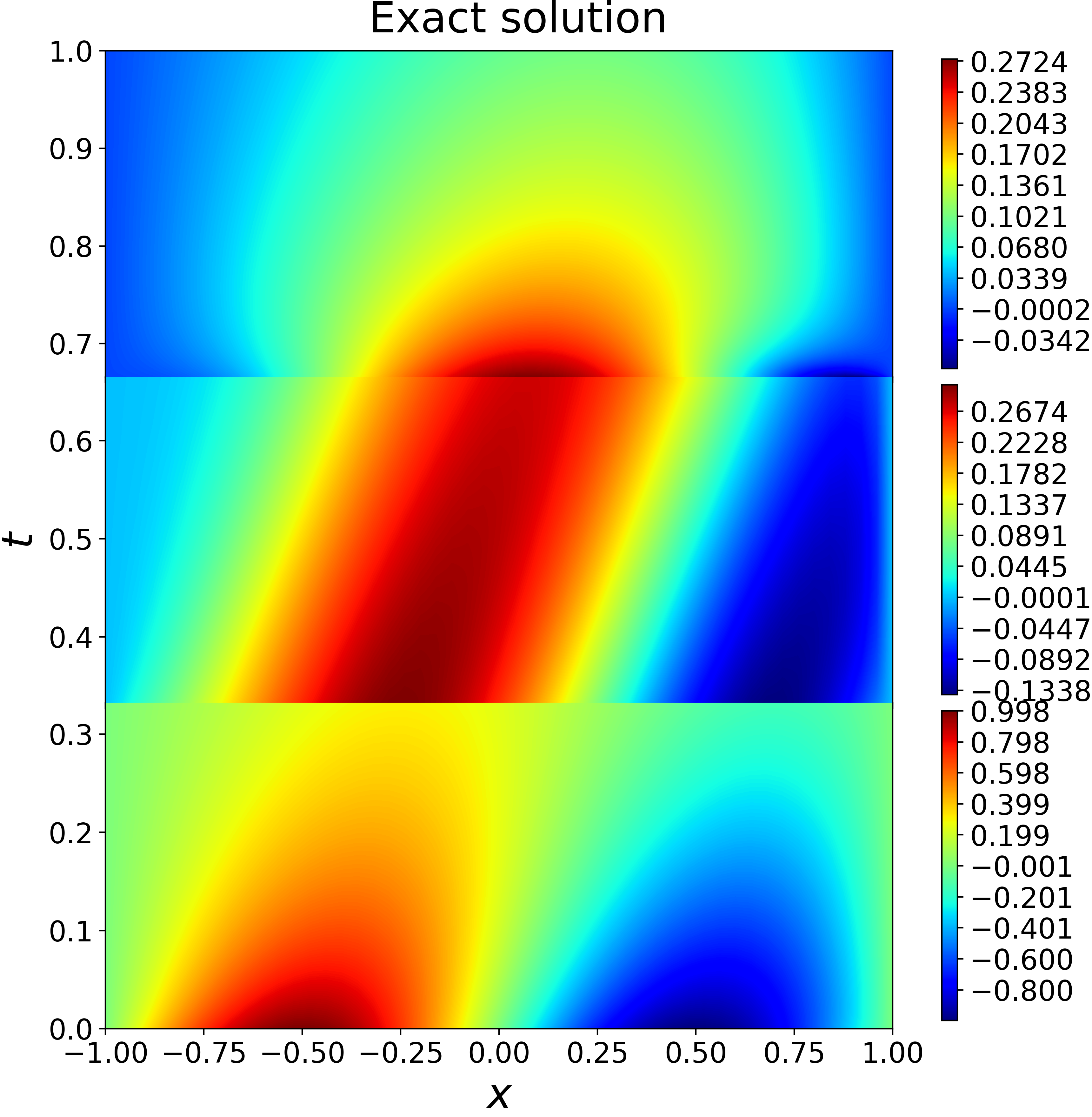}%
}\hfill
{%
  \includegraphics[height=4.5cm,width=.25\linewidth]{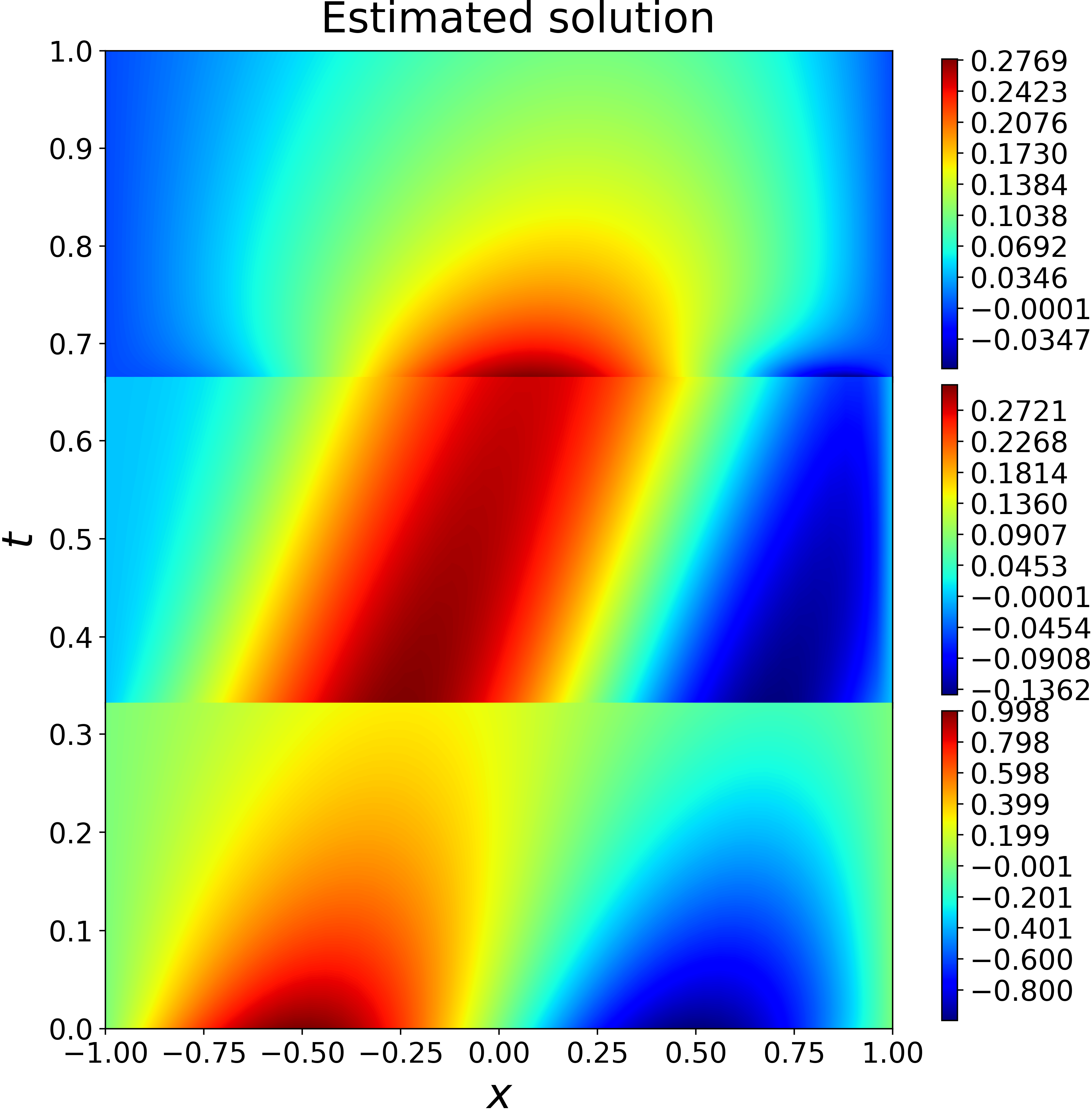}%
}\hfill
{%
  \includegraphics[height=4.5cm,width=.25\linewidth]{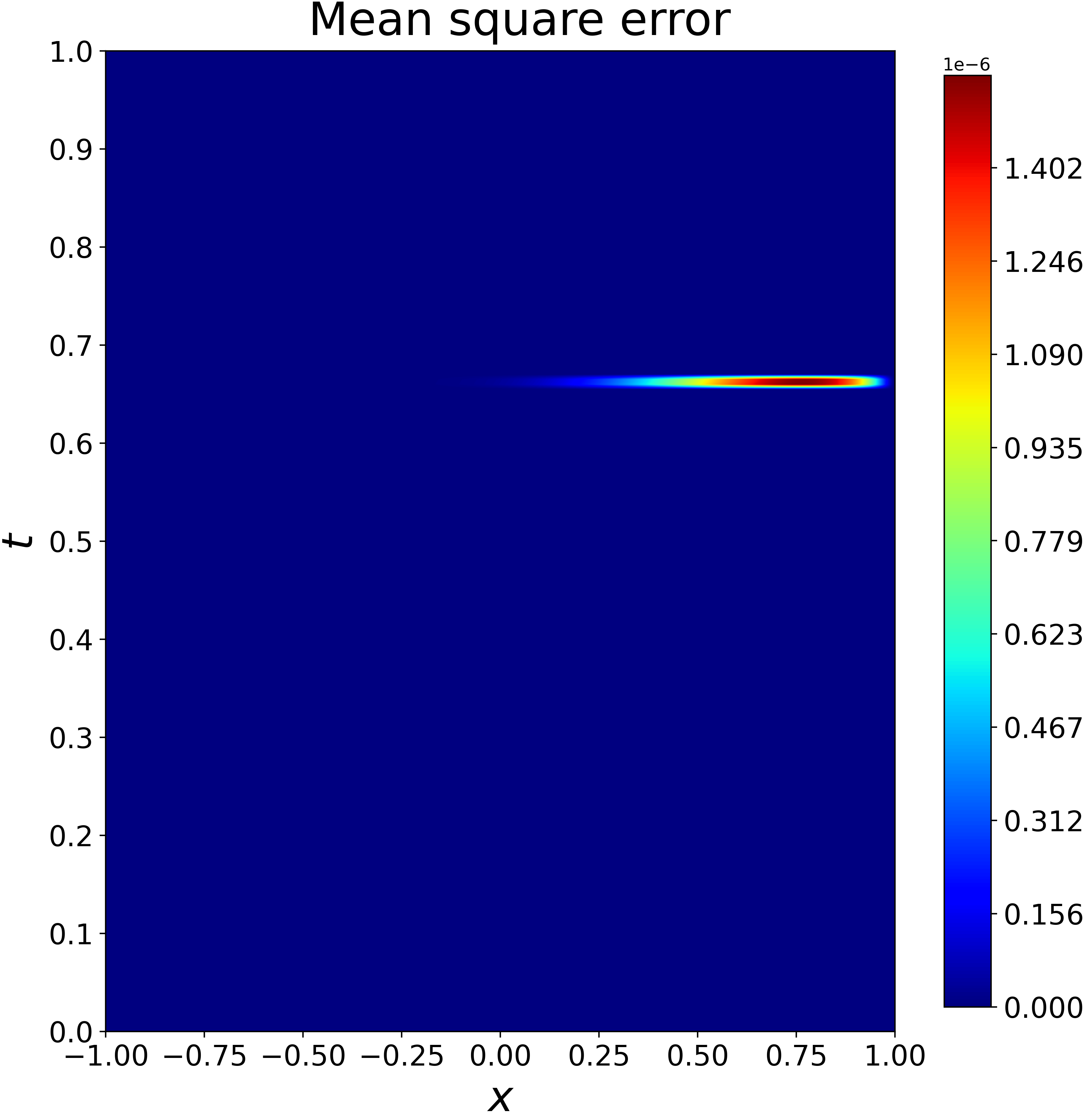}%
}
\caption{The solution of 1D advection-equation with two breakpoints.}
\label{fig: 2bp_balanced_sol}
\end{figure*}
The left panel of Figure~\ref{fig:2_change_point_ade} presents the detected changepoints over the temporal axis. In comparison, the PINNs\footnote{Standard PINNs assume a constant coefficient over time, which performs much worse for changepoint scenarios. To ensure a fair comparison, we modify PINNs to allow for a time-dependent coefficient. This highlights the effectiveness of our total variation term combined with the weights update method.} method detects these breakpoints at 0.240s and 0.615s and estimates parameters as 0.498, -0.082 and 9.018 respectively, whereas the CP-PINNs method accurately detects them at 0.334s and 0.670s and estimates parameters as 0.5001, 0.049 and 0.999 respectively. 
Figure~\ref{fig:2_change_point_ade} (right panel) illustrates the evolution of the loss function weights   $\mathbf{w}$ employing the exponentiated gradient descent method. Initially, $w_3$ marginally exceeds the other weights. Hence, the model prioritizes the detection of changepoints in the dataset. Subsequently, as the training progresses, all three weights converge to a comparable magnitude and stabilize there. This convergence indicates that the model has reached an optimal solution; the loss function has minimized to the extent that further differentiation among the weights is negligible, leading the update algorithm to allocate nearly equal adjustments to each weight term.

Figure~\ref{fig: 2bp_balanced_sol} shows the performance of CP-PINNs in discovering changepoints and solving \eqref{eq:advection}. Specifically, the leftmost panel illustrates the precise solution across a uniform temporal scale. Identifying the locations of changepoints remains challenging even when the solutions are known. In the second panel, the identical solution is depicted, but with varying scales applied to each interval between changepoints. This approach facilitates the visualization of PDEs solutions as distinct entities based on the locations of changepoints. Typically, the locations of these changepoints are not predetermined. The methodology's effectiveness in accurately identifying the locations of changepoints and providing precise solution estimations is illustrated in the third panel. Furthermore, the mean square error between the estimated and exact solutions is displayed in the fourth panel, offering a quantitative measure of the estimation accuracy.

\begin{figure*}[t]
\centering
  {%
  \includegraphics[height=4.5cm,width=.49\linewidth]{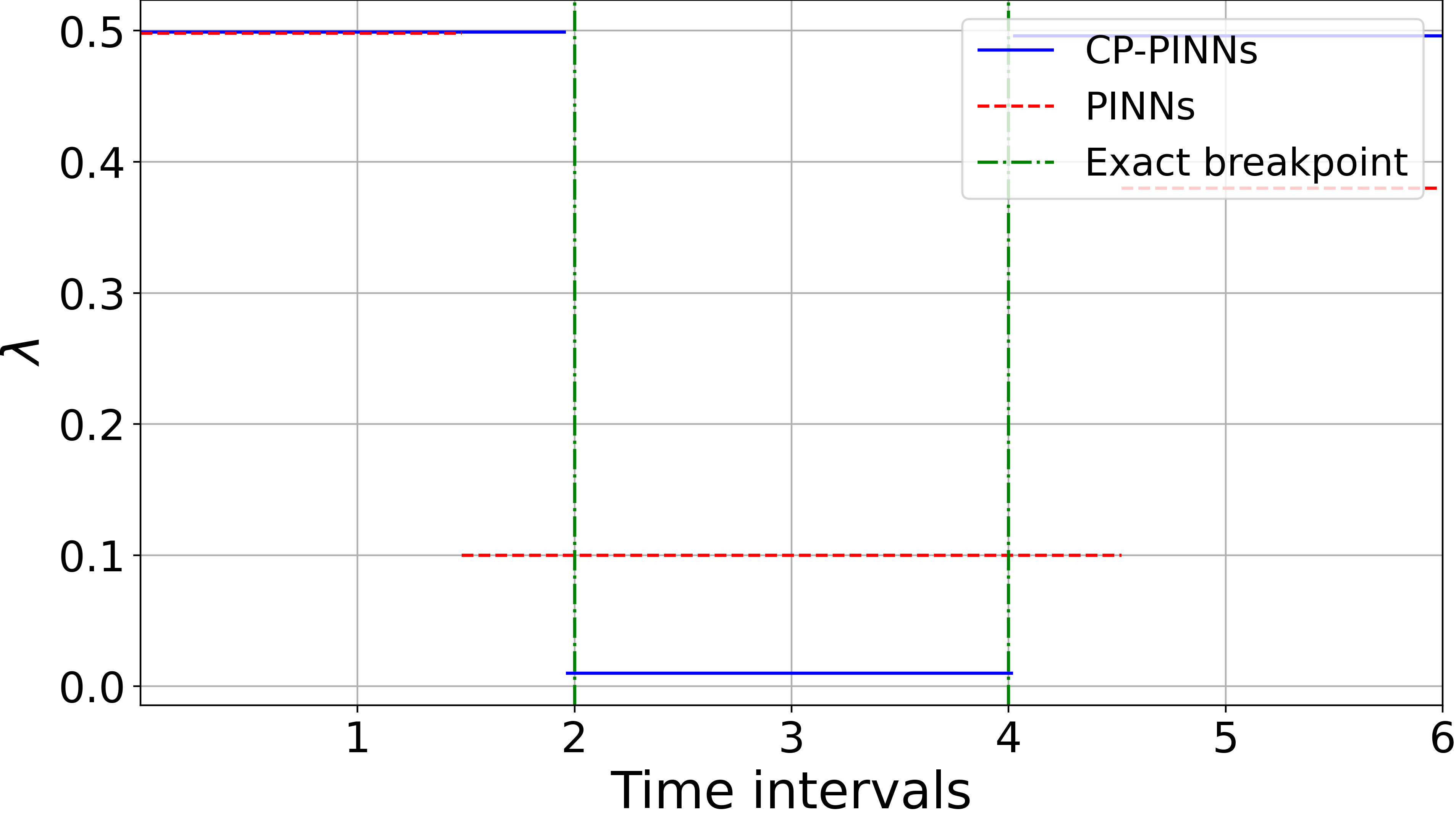}
}
  \hfill
  {%
  \includegraphics[height=4.5cm,width=.49\linewidth]{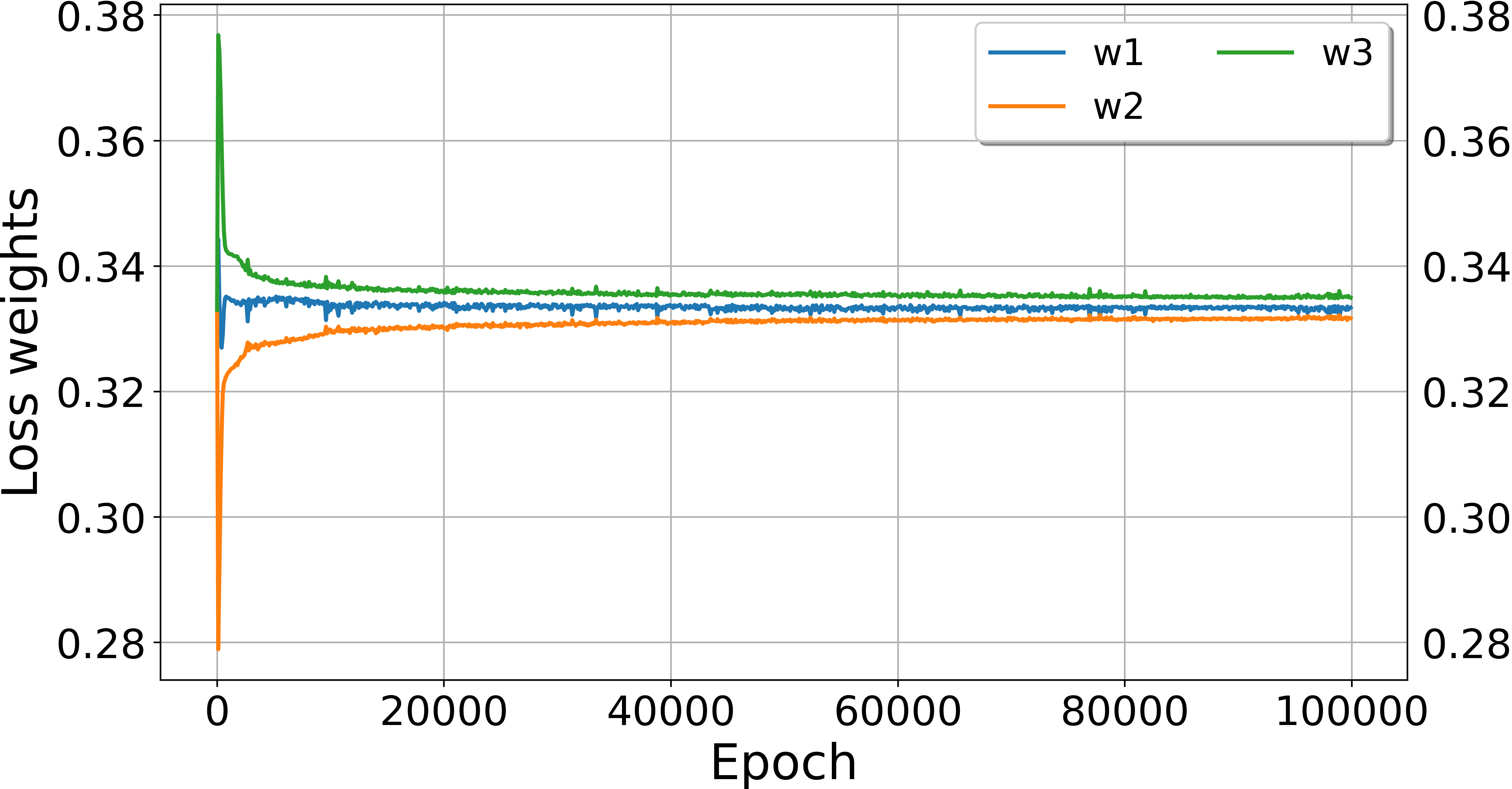}
}
\caption{(Left) Parameter estimation of CP-PINNs and PINNs. (Right) Loss weights distribution across time.}
\label{fig:ns_2_change_point_ade}
\end{figure*}

\begin{figure*}[!h]
\centering
{%
  \includegraphics[height=4.5cm,width=.33\linewidth]{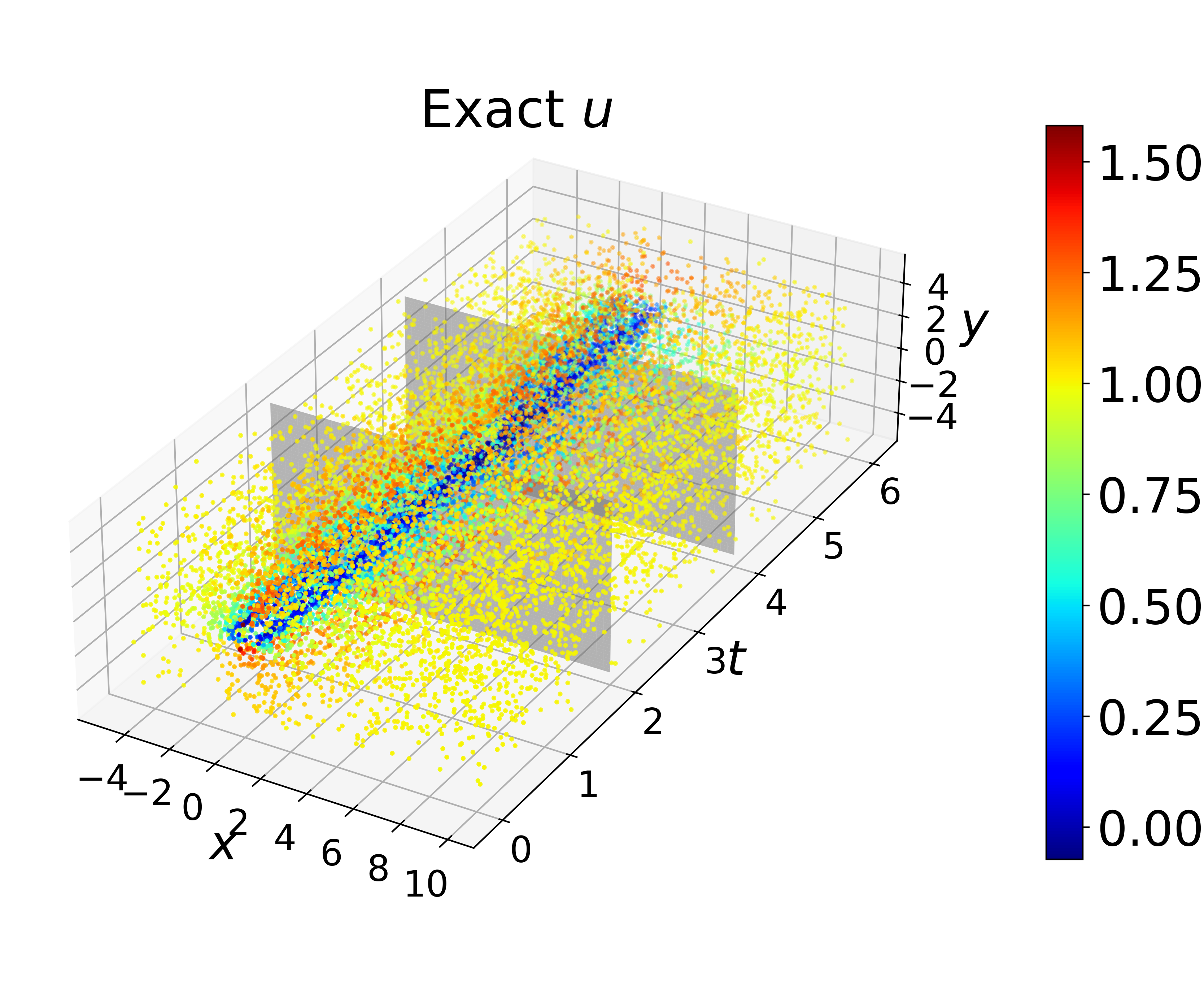}%
}\hfill
{%
  \includegraphics[height=4.5cm,width=.33\linewidth]{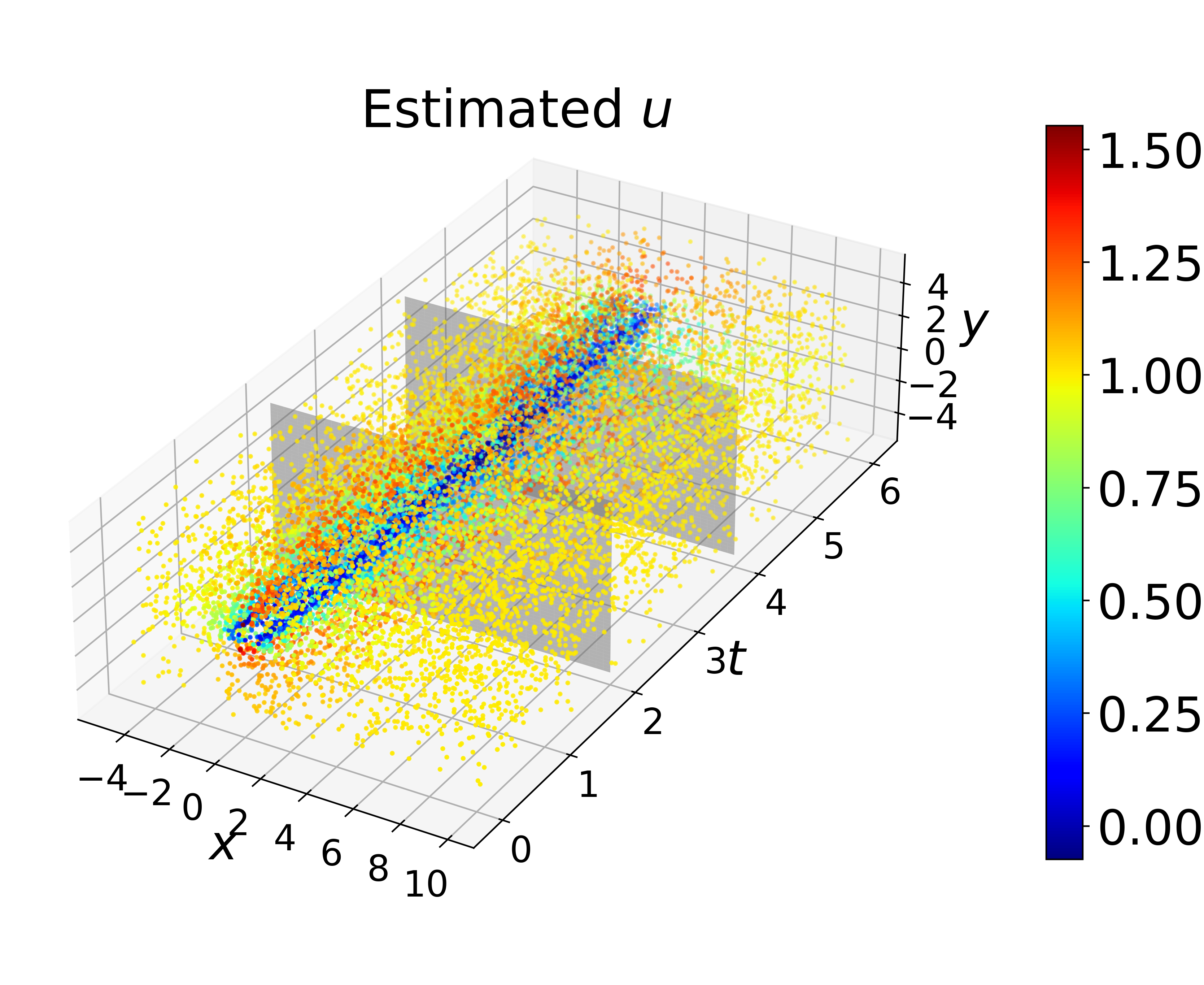}%
}\hfill
{%
  \includegraphics[height=4.5cm,width=.33\linewidth]{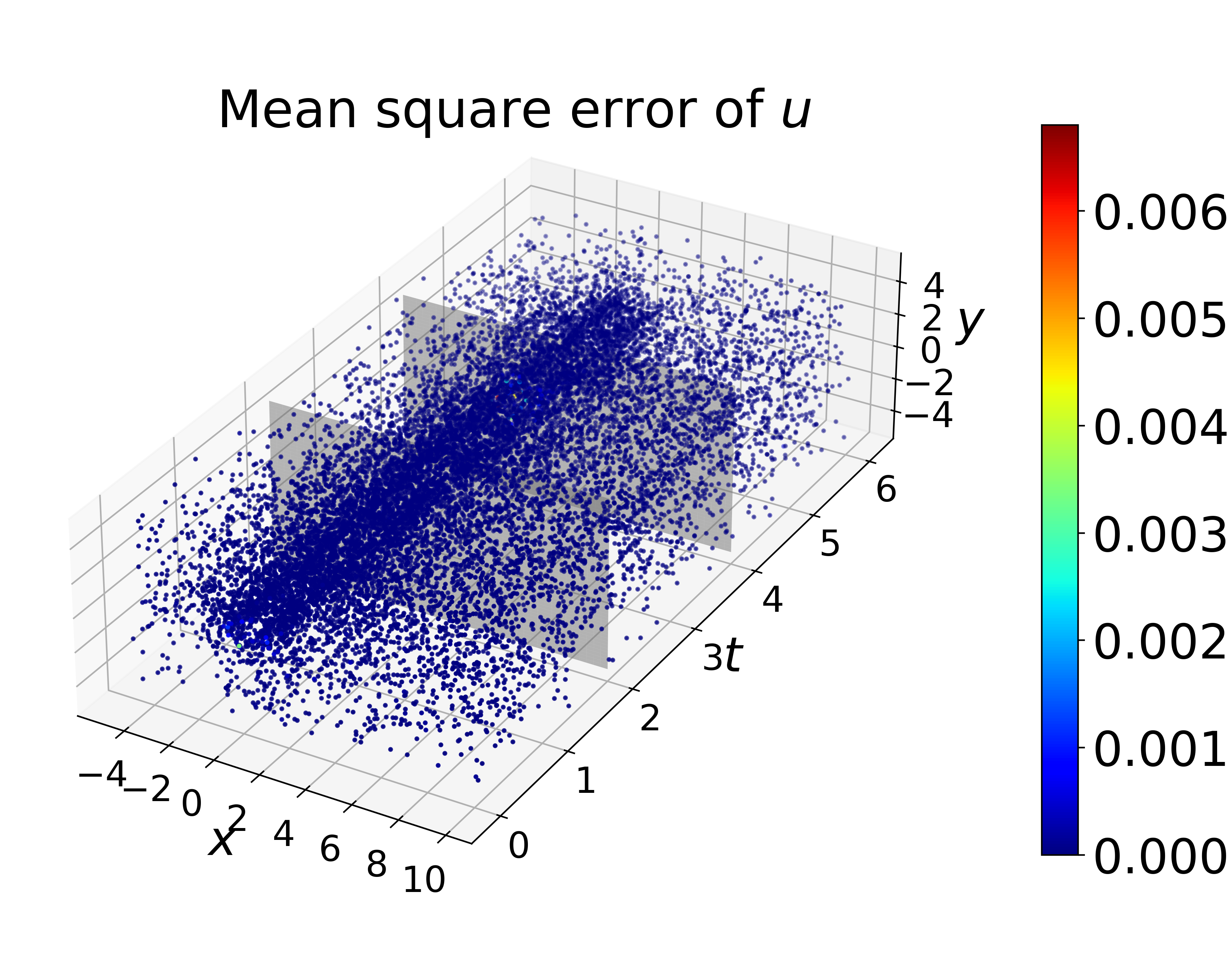}%
}
\medskip
{%
  \includegraphics[height=4.5cm,width=.33\linewidth]{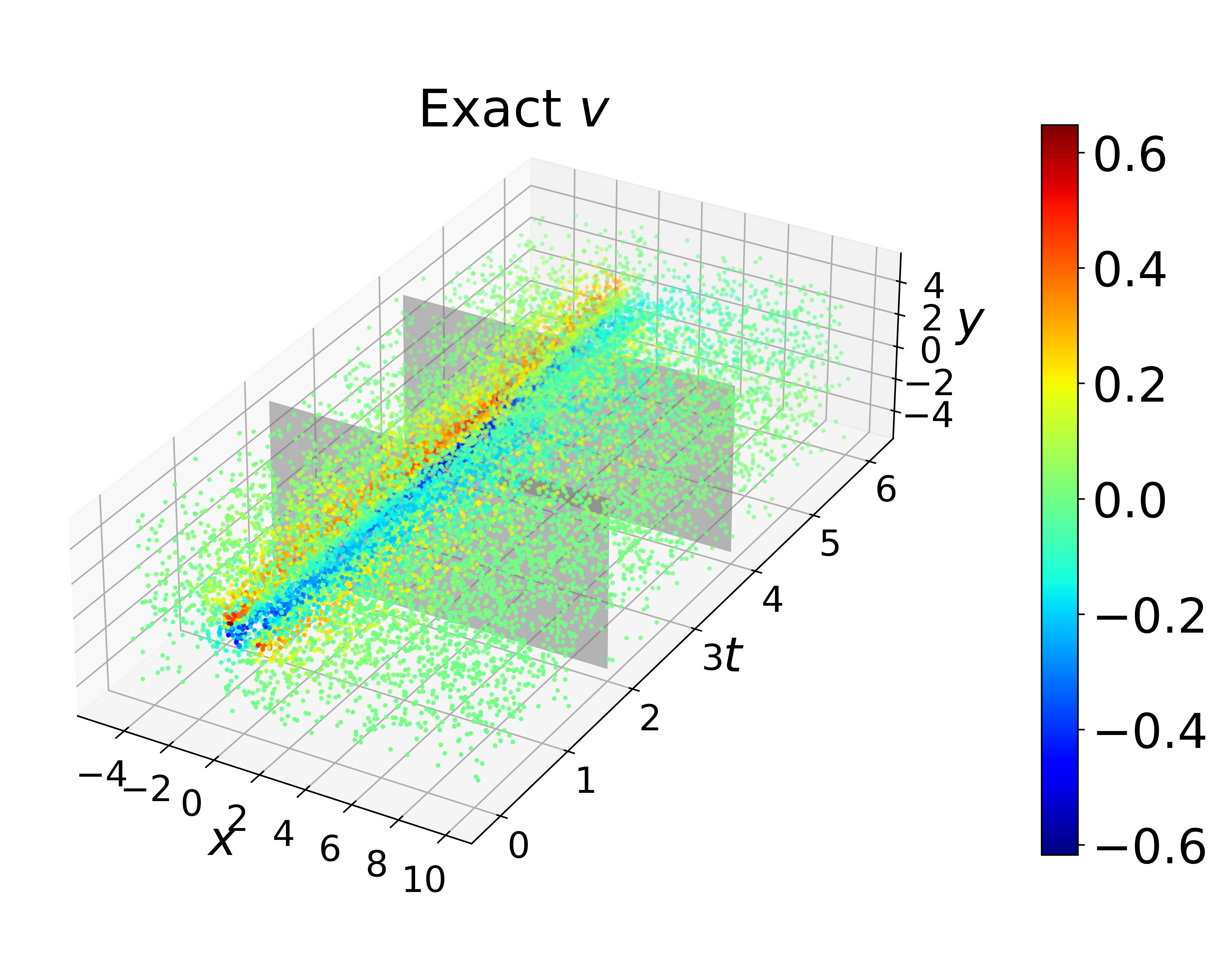}%
}\hfill
{%
  \includegraphics[height=4.5cm,width=.33\linewidth]{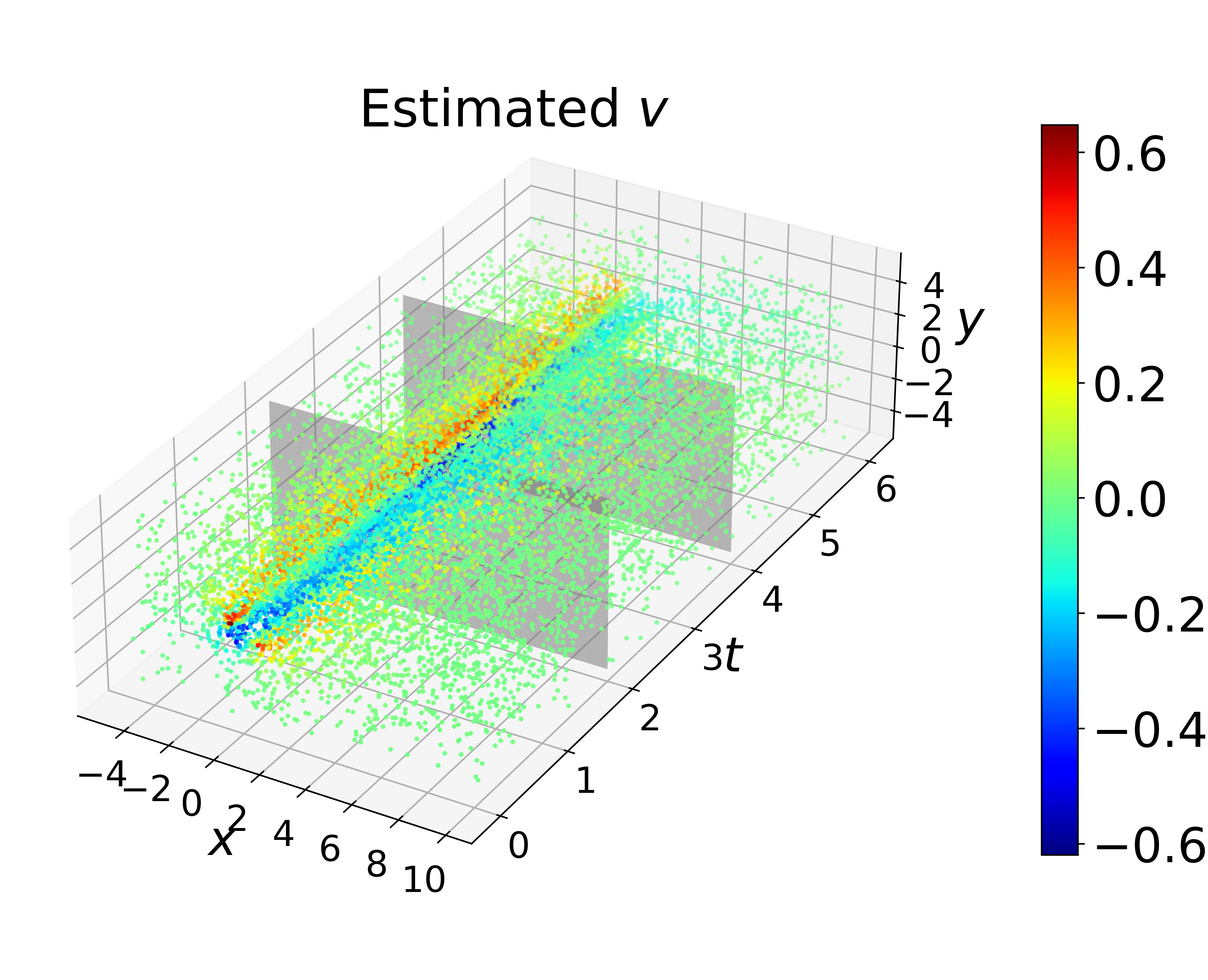}%
}\hfill
{%
  \includegraphics[height=4.5cm,width=.33\linewidth]{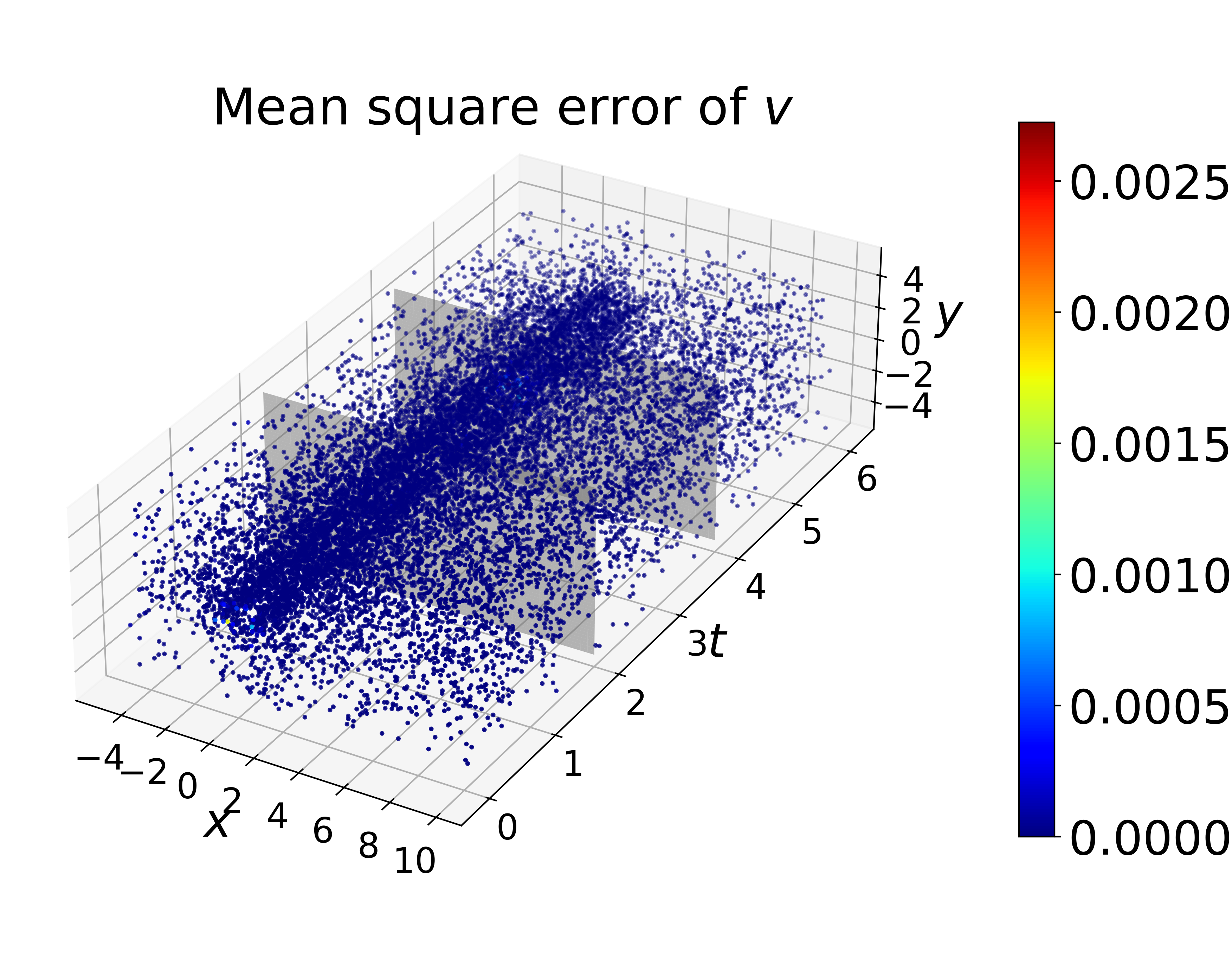}%
}
\caption{The stream-wise and transverse velocity of 2D Navier-Stokes equation with two breakpoints.}
\label{fig:ns_v_2bp_balanced_sol}
\end{figure*}
We next explore a scenario involving the Navier–Stokes equation, a standard model for incompressible fluid flow, often applied in contexts such as water flow in pipeline \cite{li2010cfd}, air dynamics over aircraft surfaces \cite{bourguet2009capturing}, and pollutant dispersion \cite{xia2001pollutant}. Here, we focus on a specialized 2D case of the equation.
\begin{equation}
    \begin{aligned}
    u_{t} + \left(u u_{x} + v u_{y}\right) &= -p_{x} + \lambda(t)\left(u_{x x} + u_{y y}\right), \\
    v_{t} + \left(u v_{x} + v v_{y}\right) &= -p_{y} + \lambda(t)\left(v_{x x} + v_{y y}\right).
\end{aligned}
\end{equation}
For $u(x, y, t)$, $v(x, y, t)$ and $p(x,y,t): \Omega \rightarrow \mathbb{R}$, where $\Omega=[-5,10] \times[-5,5] \times [0,6]$. $u(x, y, t)$ and $v(x, y, t)$  denote the stream-wise and transverse velocity components, respectively, and $p(x, y, t)$ is the measurement of fluid pressure. Similar to \cite{raissi2019physics}, we assume a uniform free stream velocity at the left boundary and a zero pressure outflow condition at the right boundary. The top and bottom boundaries are treated as periodic, ensuring continuous flow in the vertical direction. In our experiment, we focus on incompressible fluid flow passing through the 2D circular cylinder with radius = 0.5, and $\lambda(t)$ is the viscosity coefficient. It's known that fluid viscosity is significantly affected by temperature changes; for instance, heating oil decreases its viscosity, facilitating flow, while cooling increases it. Recognizing this sensitivity, we model $\lambda(t)$ to vary over time: $\lambda(t)=0.5$ for $t\in[0,2)$, then $0.01$ for $t\in[2,4)$, and returns to $0.5$ for $t\in[4,6]$. This variation simulates the effect of rapid temperature changes on fluid viscosity. 

Our method detects changepoints at 1.96s and 4.02s, offering precise parameter estimates for each segment: $\hat{\lambda}(t)=0.499$ for $t\in[0,1.96)$, $0.010$ for $t\in[1.97,4.01)$, and $0.496$ for $t\in[4.01,6]$. PINNs incorrectly estimate changepoints at 1.48s and 4.52s, yielding the erroneous parameters 0.498, 0.101, and 0.38, respectively. Figure~\ref{fig:ns_v_2bp_balanced_sol} showcases our model's capability in solving both stream-wise and traverse velocities. The left panel of Figure~\ref{fig:ns_2_change_point_ade} presents the parameter estimation when changepoints are present. Analogous to Figure~\ref{fig:2_change_point_ade}, the right panel demonstrates that upon identifying the optimal estimates, the three weights converge to approximately equal values.
\section{Conclusions}\label{sec:Conclusion}
We introduce a novel method for identifying changepoints in dynamic systems governed by general PDEs dynamics. Our approach works with piecewise-constant time-changing parameters and leverages total variation regularization on the first-order differences of parameters. We also propose an online learning strategy that balances the priorities between changepoints detection, model fitting, and PDEs discovery. For future works, we plan to extend our research to more complex scenarios, including PDEs with parameters that are arbitrary functions of the time domain. Additionally, we seek to investigate changepoints that are associated with both temporal and spatial variations, such as Quasi-linear PDEs.

\section*{Appendix}
The following are proofs of Lemma \ref{lemma:update_weigts} and Corollary \ref{cor:regret}.
\begin{proof}
From the linearity of the loss function  $L(\mathbf{w})$ we can expand it around $\mathbf{w}^{(k-1)}$, i.e., 
\begin{equation}\label{eq:taylor}
L(\mathbf{w}) = L(\mathbf{w}^{(k-1)}) + \nabla_w L(\mathbf{w}^{(k-1)})^\top \left( \mathbf{w} - \mathbf{w}^{(k-1)}\right).
\end{equation}
We choose $\frac{1}{\eta} \mathbf{w}^\top \log \mathbf{w}$ as the regularization, where $\eta>0$ is the learning rate. 

Since $L(\mathbf{w}^{(k-1)})$ and $\nabla_w L(\mathbf{w}^{(k-1)})^\top\mathbf{w}^{(k-1)}$ are independent of $\mathbf{w}$, we can drop these terms and use

the Lagrange multiplier and the simplex definition for some fixed $\gamma$, we have
\begin{equation}
\begin{aligned}
\mathbf{w}^{(k)} = &\underset{\mathbf{w},\gamma}{\operatorname{argmin}} \left\{\nabla_w L(\mathbf{w}^{(k-1)})^\top \mathbf{w}  + \frac{1}{\eta} \mathbf{w}^\top \log \mathbf{w} + \gamma (1- \mathbf{w}^\top \mathbf{1}_3  )\right\}.
\end{aligned}
\end{equation}
We take the derivatives with respect to $\mathbf{w}$ and set it to zero, 
\begin{equation}
    \nabla_w L(\mathbf{w}^{(k-1)}) + \frac{1}{\eta} (\mathbf{1}_3 + \log \mathbf{w}) - \gamma\mathbf{1}_3 = \mathbf{0}
\end{equation}
We can get 
     $\mathbf{w}^{(k)} = \exp\left[ -\eta\nabla_w L(\mathbf{w}^{(k-1)}) - (\mathbf{1}_3-\eta\gamma\mathbf{1}_3)\right]$ and $\gamma=\frac{1}{\eta}\left[1-\log \left(\exp\left(-\eta\nabla_w L(\mathbf{w}^{(k-1)}) \right)^\top\mathbf{1}_3\right)\right]$.
\end{proof}
\begin{proof}
    First, we show $\frac{1}{\eta} \mathbf{w}^\top \log \mathbf{w}$ is a strong convex regularizer.
    
    For $\forall \mathbf{u}, \mathbf{w} \in [0,1]^3$, we consider the Kullback-Leibler divergence $\mathbf{u}^\top (\log \mathbf{u} - \log \mathbf{w})$. By Pinsker’s Inequality, we have
    \begin{equation}
      \mathbf{u}^\top (\log \mathbf{u} - \log \mathbf{w}) \geq 2\|\mathbf{u}-\mathbf{w}\|_1^2 \geq \frac{1}{2}\|\mathbf{u}-\mathbf{w}\|_1^2.
    \end{equation}
    Adding $ \mathbf{u}^\top \log \mathbf{w}$ from both sides, we have
    \begin{equation}
    \begin{aligned}
        &\ \mathbf{u}^\top \log \mathbf{u} \geq  \mathbf{u}^\top \log \mathbf{w} + \frac{1}{2}\|\mathbf{u}-\mathbf{w}\|_1^2 \\
        &= \mathbf{w}^\top\log\mathbf{w} + \left(\mathbf{u} -\mathbf{w}\right)^\top\left(\mathbf{1}_3+\log \mathbf{w}\right) + \frac{1}{2}\|\mathbf{u}-\mathbf{w}\|_1^2.
    \end{aligned}
    \end{equation}
    Then we have
    \begin{equation}
        \frac{1}{\eta}\mathbf{u}^\top \log \mathbf{u} \geq  \frac{1}{\eta}\mathbf{w}^\top\log\mathbf{w} + \frac{1}{\eta}\left(\mathbf{u} -\mathbf{w}\right)^\top\left(\mathbf{1}_3+\log \mathbf{w}\right) + \frac{1}{2\eta}\|\mathbf{u}-\mathbf{w}\|_1^2.
    \end{equation}
    This shows the regularizer is $\frac{1}{\eta}$ strong convexity.
    We consider the first order Taylor Expansion, the loss function $L(\mathbf{w}^{*})$ is a linear function of $\mathbf{w}^{*}$, where $\forall \mathbf{w}^{*}$ with $\mathbf{w}^{*}\in \mathcal{S}_3$, and $\|\nabla_w L(\mathbf{w}^{(k-1)})\|_1 \leq G$, Lemma 2.3 in \cite{shalev2012online} implies that
\begin{equation}
\begin{aligned}
\operatorname{Regret} &\leq \sum_{k=1}^{B}\left(\nabla_w L(\mathbf{w}^{(k-1)}) - \nabla_w L(\mathbf{w}^{(k)})\right)^\top\mathbf{1}_3  \\
&+ \frac{1}{\eta}(\mathbf{w}^{*})^\top \log \mathbf{w}^{*} - \frac{1}{\eta}(\mathbf{w}^{(1)})^\top \log \mathbf{w}^{(1)}.    
\end{aligned}
\end{equation}
Since the regularizer is $\frac{1}{\eta}$ strong convexity, using Theorem 2.10 in \cite{shalev2012online}, we have 
\begin{equation}
\begin{aligned}
    \operatorname{Regret} &\leq \frac{1}{\eta}(\mathbf{w}^{*})^\top \log \mathbf{w}^{*} - \frac{1}{\eta}(\mathbf{w}^{(1)})^\top \log \mathbf{w}^{(1)} + \eta B G^2 \\
    &\leq \frac{\log 3}{\eta} + \eta B G^2,
\end{aligned}
\end{equation}
because $-\log 3 \leq \mathbf{w}^\top \log \mathbf{w} \leq 0$, for $\mathbf{w} \in \mathcal{S}_3$.
\end{proof}

%Bibliography
\bibliographystyle{unsrt}  
\bibliography{references}  
\end{document}